\documentclass[sigconf, screen, authorversion]{acmart}
\AtBeginDocument{%
  \providecommand\BibTeX{{%
    \normalfont B\kern-0.5em{\scshape i\kern-0.25em b}\kern-0.8em\TeX}}}

\usepackage{bm}
\usepackage{multirow}
\usepackage{epsfig}
\usepackage{subfig}
\usepackage{subcaption}
\usepackage{graphicx}
\usepackage{pifont}
\usepackage{xcolor}
\usepackage{tabularx}
\usepackage{colortbl}
\usepackage{svg}
\usepackage{caption}
\usepackage{enumitem}

\newcommand{\etal}{\textit{et al.}}

\setcopyright{acmlicensed}
\copyrightyear{2024}
\acmYear{2024}
\setcopyright{acmlicensed}\acmConference[ICMI '24]{INTERNATIONAL CONFERENCE ON MULTIMODAL INTERACTION}{November 4--8, 2024}{San Jose, Costa Rica}
\acmBooktitle{INTERNATIONAL CONFERENCE ON MULTIMODAL INTERACTION (ICMI '24), November 4--8, 2024, San Jose, Costa Rica}
\acmDOI{10.1145/3678957.3685707}
\acmISBN{979-8-4007-0462-8/24/11}

\begin{document}

\title[Learning Co-Speech Gesture Representations in Dialogue through Contrastive Learning]{Learning Co-Speech Gesture Representations in Dialogue through Contrastive Learning: An Intrinsic Evaluation}

\author{Esam Ghaleb}
\orcid{0000-0002-0603-9817}
\affiliation{%
  \institution{University of Amsterdam}
  \streetaddress{Science Park 904}
  \city{Amsterdam}
  \country{The Netherlands}
  \postcode{1098 XH}
}
\email{e.ghaleb@uva.nl}

\author{Bulat Khaertdinov}
\affiliation{%
  \institution{Maastricht University}
  \country{The Netherlands}
}

\author{Wim Pouw}
\affiliation{%
  \institution{Radboud University}
  \country{The Netherlands}
}

\author{Marlou Rasenberg}
\affiliation{%
  \institution{Meertens Institute}
  \country{The Netherlands}
}

\author{Judith Holler \& Asl\i~Özy\"{u}rek}
\affiliation{%
  \institution{Radboud University \& MPI for Psycholinguistics}
  \country{The Netherlands}
}

\author{Raquel Fern\'{a}ndez}
\orcid{0000-0001-5540-5943}
\affiliation{%
  \institution{University of Amsterdam}
  \country{The Netherlands}
}
\email{raquel.fernandez@uva.nl}

\renewcommand{\shortauthors}{Ghaleb et al.}

\begin{abstract}
In face-to-face dialogues, the form-meaning relationship of co-speech gestures varies depending on contextual factors such as what the gestures refer to and the individual characteristics of speakers. These factors make co-speech gesture representation learning challenging. How can we learn meaningful gestures representations considering gestures' variability and relationship with speech? This paper tackles this challenge by employing self-supervised contrastive learning techniques to learn gesture representations from skeletal and speech information. We propose an approach that includes both unimodal and multimodal pre-training to ground gesture representations in co-occurring speech. For training, we utilize a face-to-face dialogue dataset rich with representational iconic gestures. We conduct thorough intrinsic evaluations of the learned representations through comparison with human-annotated pairwise gesture similarity. Moreover, we perform a diagnostic probing analysis to assess the possibility of recovering interpretable gesture features from the learned representations. Our results show a significant positive correlation with human-annotated gesture similarity and reveal that the similarity between the learned representations is consistent with well-motivated patterns related to the dynamics of dialogue interaction. Moreover, our findings demonstrate that several features concerning the form of gestures can be recovered from the latent representations. Overall, this study shows that multimodal contrastive learning is a promising approach for learning gesture representations, which opens the door to using such representations in larger-scale gesture analysis studies.
\end{abstract}



\received{05 May 2024}
\received[accepted]{18 July 2024}
\maketitle

\section{Introduction}
\label{sect:intro}
Co-speech hand gestures 
are intentionally used along with speech to convey meaning \cite{mcneill1992hand}. 
For instance, representational iconic gestures depict objects, events or actions through various representational techniques such as enacting, tracing, and hand-shaping \cite{kendon2004gesture}. Gesture analysis is an active research area in fields such as Human-Computer Interaction (HCI) \cite{Liu2022b}, Sign Language Recognition (SLR) \cite{kong2014towards, jiang2021skeleton}, and human behavior analysis \cite{kucherenko2019analyzing, eijk2022cabb},  
where sensory data collected through wearable sensors \cite{guo2021human} or, more commonly, through passive sensors like RGB or depth cameras are widely used for studying gestures \cite{pouw2020quantification, trettenbrein2021controlling,trujillo2019toward}. 

In face-to-face interaction, the form-meaning relationship of co-speech gestures is influenced by various situational and contextual factors, including what a gesture refers to and the characteristics of individual speakers.
Although multiple current studies aim to model and represent gestures, there are prominent areas with room for improvement, particularly concerning gesture representation learning in conversations~\cite{wang2017large, ghaleb2023co, ghaleb2024le, xu2024chain, yoon2020speech, liu9beat}.
\textit{First}, most studies train deep learning architectures from scratch on specific downstream tasks, including gesture segmentation \cite{wang2017large, ghaleb2023co, ghaleb2024le} or generation \cite{xu2024chain, yoon2020speech, liu9beat}. Thus, the employed objectives are focused on the task-specific discriminative and generative power of the models rather than on their ability to effectively encode general meaningful properties of gestures and relationships between them.
The research literature has already pointed out the lack of models to represent gestures \cite{yoon2020speech, liu9beat, xu2024chain}. For example, some authors use autoencoders to extract and compare latent representations to evaluate synthetic gestures \cite{yoon2020speech, liu9beat, xu2024chain}.
\textit{Second}, much of the research has also focused on emblems (which are conventionalized and can stand alone without speech) \cite{zhu2018continuous, kopuklu2019real, benitez2021ipn} or pure beat gestures (rhythmic gestures that lack semantic content), particularly within monologue-based datasets as surveyed in \cite{Nyatsanga2023}.
Yet, in real-world interactions, gestures are semantically, pragmatically, and temporally related to speech, especially in face-to-face conversations. For instance, along with speech, representational gestures during face-to-face interaction facilitate the identification of referents
and the establishment of shared understanding \cite{holler2011co}.

In this study, we address these limitations by focusing on learning representations of co-speech gestures in face-to-face dialogue. These representations should exhibit desirable intrinsic properties, i.e., properties that align with expert intuitions.
We propose to learn such representations using self-supervised learning approaches. In particular, building upon a pre-trained sign-language recognition model~\cite{jiang2021skeleton}, we exploit unimodal (skeletal) and multimodal (speech-skeletal) contrastive learning to encode co-occurring gestural and speech patterns without using any manually annotated features. 
We opt for contrastive learning objectives because 
they allow us to simultaneously refine visual gesture representations and ground them in robust speech features learned by a foundation speech model. Furthermore, such a framework opens up possibilities for integrating other sources of information, e.g., semantic information through models processing text, in future work.

For training models with the proposed objectives, we utilize a dataset of naturalistic face-to-face \textit{referential communication} rich with representational iconic gestures collected by Rasenberg \etal~\cite{rasenberg2022primacy}. In this dataset, speakers engage in a conversation to identify objects that they are not familiar with. 
Gesture strokes (i.e., the most meaning-bearing component of a gesture \cite{mcneill1992hand}) are manually segmented by experts, 
and we take these segments as our starting point for representation learning.
Moreover, a subset of pairs of gestures by two different speakers were coded by the dataset authors as similar or dissimilar with respect to five form features: handedness, position, shape, rotation, and movement. We use this annotation to evaluate the intrinsic quality of the gesture representations learned by our models. 
To summarise, we make the following contributions:

\begin{itemize}[leftmargin=10pt]
    \item We propose a self-supervised contrastive learning approach to simultaneously learn co-speech gesture representations through skeletal information while grounding them in spoken language.
    \item We present extensive intrinsic evaluations of the learned representations by investigating how well they capture pairwise gesture similarity. Our results show a significant positive correlation with human-annotated form feature similarity and reveal that the similarity between the learned representations is consistent with well-motivated patterns concerning speakers, referents, and the dynamics of dialogue interaction.
    \item Finally, we conduct a diagnostic probing analysis to investigate whether the model representations implicitly encode interpretable gesture features, finding that several form features can be recovered to some extent from the latent representations learned via self-supervised contrastive learning.
\end{itemize}

\noindent
Overall, this study shows that multimodal contrastive learning is a promising approach to learning gesture representations whose properties are well aligned with human judgments and with theoretically motivated expectations. This paves the way for using self-supervised gesture representations to scale up gesture analysis studies that currently rely on small hand-annotated samples.

\section{Related Work}\label{sect:related_work}
\subsection{Gesture Representation and Modelling}
\subsubsection{Gesture Representation and Similarity}
Prior to the rise of deep learning, researchers utilized handcrafted features, such as velocity, rhythm, acceleration, and anatomical models of hands, to encode kinematic, physical, and shape characteristics of gestures \cite{rautaray2015vision}. Nowadays, the focus has shifted to training neural networks end-to-end for certain downstream tasks, such as gesture recognition \cite{liu20203d}, detection \cite{ghaleb2024le}, and generation \cite{yoon2020speech}. 
The studies exploring gesture generation frequently involve quantitative evaluations for the quality and fidelity of synthetic gestures using custom metrics, such as Fréchet Gesture Distance (FGD) and Beat Alignment Score (BeatAlign) \cite{yoon2020speech, xu2024chain}. For example, FGD evaluates generated gestures based on learned visual features through autoencoders \cite{yoon2020speech, Liu2022a}.

The similarity of gesture or sign pairs has been studied in a range of contexts, for example, in silent gestures, co-speech gestures, co-singing gestures, as well as sign language generation and production \cite{kanakanti2023multifacet}. 
For instance, Kanakanti \etal~\cite{kanakanti2023multifacet} used Dynamic Time Warping (DTW) to evaluate speech-to-sign language generation models. These metrics assess the alignment between the predicted sign language sequences and the ground truth of Indian Sign Language sequences. 
One drawback of spatio-temporal alignment metrics such as DTW is that they operate at the level of key points and might not be optimal for measuring the more context-dependent similarity of two co-speech gestures. Indeed, a very weak but reliable correlation is generally found between the dissimilarity of gesture kinematics relative to the dissimilarity of what the gestures are about \cite{pouw2021semantically}. In our work, we assess the properties of the learned representations by performing pairwise gestural analysis. Hence, FGD and BeatAlign are not suitable metrics for this study. 
Rather, we calculate the cosine similarity of the learned representations, as often done in Natural Language Processing 
and speech research, where it is common to use embeddings 
for representational and similarity analysis \cite{navigli2019overview, bruni2014multimodal, pezzelle-etal-2021-tacl, pasad2024self}.

\subsubsection{Joint Speech and Gesture Models}
Joint modeling of speech and gestures through speech and vision models has been studied through different gesture analysis tasks. For instance, Ghaleb et al.~\cite {ghaleb2024le} used speech to enhance gesture detection methods. A study by Kucherenko et al. \cite{kucherenko2022multimodal} found that the meaning and timing of gestures can be predicted using prosodically relevant acoustic 
features. %
In gesture generation, according to a survey by Nyatsanga \etal~\cite{Nyatsanga2023}, adding information about speech prosody and meaning can improve the quality of generated beat and representational gestures, respectively, making them appear more natural based on evaluations by raters. A typical pipeline representing gesture generation models is the work by Bhattacharya \etal~\cite{bhattacharya2021speech2affectivegestures}, who developed a dual-speech model. The first model uses prosodic features, while the second one leverages word embeddings to capture speech's prosodic and semantic characteristics separately.
Another example closely related to ours is the study by Lee \etal~\cite{lee2021crossmodal}, who proposed an approach that exploits spoken language and gestures for gesture generation through cross-modal contrastive learning. The authors cluster speech and gesture embeddings cross-modally. In contrast, we include cross-modal objectives to complement unimodal gesture representations, as explained in Section~\ref{sect:models}.

In sum, gesture representations are typically learned through supervision signals tailored for a certain task. %
Such an approach might be suboptimal for learning general-purpose gesture representations. %
Autoencoders, in turn, are an established %
technique for this purpose in unimodal settings \cite{yoon2020speech, liu9beat}, 
but are not frequently used when researchers are interested in constructing \textit{multimodal} representations. 
In this paper, we aim to learn general gesture representations that capture properties that align with expert intuitions, grounding their visual form-based characteristics in speech signals via unimodal and cross-modal contrastive learning.

\subsection{Pre-training through Contrastive Learning}
Contrastive learning has emerged as an efficient self-supervised learning strategy for pre-training deep learning models across various fields without using data annotations \cite{radford2021learning, jaiswal2020survey}. The main idea behind this paradigm is to maximize a similarity metric computed between pairs of semantically close instances, i.e., positive pairs, by contrasting them against semantically different or negative pairs. The contrastive learning approach can be flexibly used for both unimodal \cite{chen2020simple, he2020momentum} and multimodal \cite{tian2020contrastive, girdhar2023imagebind} representation learning. 

Both unimodal and multimodal contrastive objectives have been applied to skeleton data. Without ground truth annotations, semantic similarity is typically defined by the correspondence of feature representations to instances. In the unimodal case, a positive pair is typically formed by two views augmented from the same instance \cite{chen2020simple}. Building upon this idea, multiple works \cite{thoker2021skeleton, guo2022contrastive} propose to use various spatial and temporal augmentations applied to skeletal key points for pre-training of graph- and sequence-based architectures.
In turn, multi-view and multimodal contrastive objectives treat an instance represented through different views and modalities as a positive pair \cite{tian2020contrastive, radford2021learning}. Applied to skeletal data, these approaches aim to pre-train models by maximizing instance-level similarities of representations extracted from different views of skeletons, e.g., joints and motion \cite{li20213d}, or modalities such as RGB and depth videos \cite{hong2022versatile} or inertial measurement units \cite{brinzea2022contrastive}. In this paper, motivated by the effectiveness of contrastive self-supervised learning, we adopt both unimodal and multimodal objectives to learn representations of co-speech gestures without using data annotations.

\section{Dataset and Pre-Processing}
\label{sect:setup}

\subsection{Dataset}
\begin{figure}[th]
    \centering
    \includegraphics[width=0.8\linewidth]{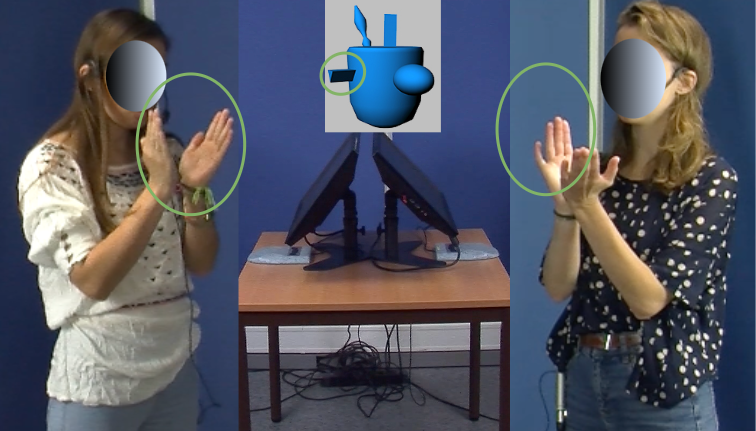}
    \caption{Pair of gestures referring to the highlighted subpart of the non-conventional object. The gesture pair is coded as similar in all form features: handedness (speakers use both hands), shape, position, rotation, and movement.}
    \label{fig:similarity_example}
    \Description[]{}

\end{figure}

We use a dataset collected by Rasenberg \etal~\cite{rasenberg2022primacy} that consists of 19 naturalistic face-to-face dialogues. In these dialogues, Dutch-speaking participants play a referential game where they need to identify different objects or \textit{referents} that do not have a conventional label. The game setup is shown in Figure \ref{fig:similarity_example}. Each pair of participants plays the game for six rounds. In each round, one speaker acts as the `director,' describing one of the target objects, while the other speaker acts as a `matcher' who attempts to identify the object among several candidates. Each round includes 16 trials (the total number of objects), and the speakers exchange the director-matcher roles after each trial. The speakers were recorded from three angles, and we use video recordings from semi-frontal views (as shown in Figure \ref{fig:similarity_example}). The dataset includes 38 subjects and more than 8 hours of recordings and is part of the larger CABB dataset \cite{eijk2022cabb}. %

The speakers are free to communicate as they please while playing the game;
they were not given any instructions on the use of gestures. However, due to the nature of the referential task, the dataset is rich with representational iconic gestures, i.e., gestures used to depict objects. In this dataset, gesture strokes were manually identified and segmented. 
In addition, gestures were annotated according to what subpart of an object they refer to (see an example object and highlighted subpart in Figure \ref{fig:similarity_example}). This annotation resulted in $4949$ gesture segments with an average duration of $610$ milliseconds.

\subsection{Manual Similarity Coding of Gesture Pairs}
\label{sect:manual_annot}

A subset of the dataset was annotated by Rasenberg \etal~\cite{rasenberg2022primacy} to study the extent to which speakers mimic each other through gestures. 
The researchers coded the similarity of $419$ semantically related pairs of gestures (i.e., gestures referring to the same subpart as shown in Figure \ref{fig:similarity_example}) made by two speakers within a dialogue. The annotation codes, in a binary manner, whether two gestures are similar or not with respect to five form features: shape, movement, rotation, position, and handedness (whether both speakers use the left hand, the right hand, or both hands). 
This set of features is well-motivated and close to the set used by Bergmann \& Kopp \cite{bergmann2012gestural}, who studied the similarity of temporally related gestures. 
As an example, the two gestures in Figure \ref{fig:similarity_example} were coded as similar for all features since both speakers use both hands with the same shape, movement, position, and rotation. 

\begin{figure}[htb]
    \includegraphics[width=0.6\linewidth]{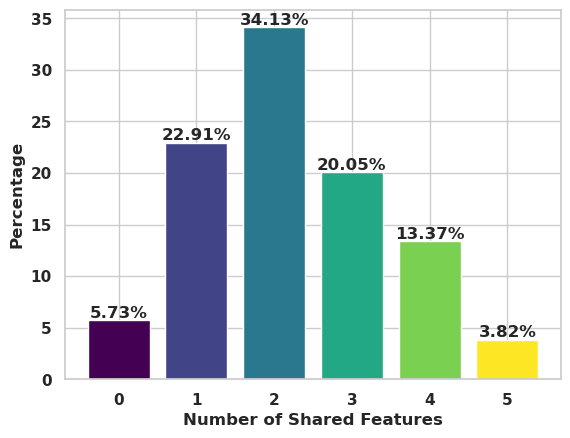}
    \caption{Distribution of the number of shared form features in the gesture pairs manually annotated by Rasenberg~\etal}
    \label{fig:distribution_of_shared_feats}
    \Description[]{}
\end{figure}

The distribution of the number of form features for which the gesture pairs are similar can be found in Figure \ref{fig:distribution_of_shared_feats}. 
The majority of referentially related gesture pairs by two speakers are similar with respect to at least one feature, and around 38\% are similar with respect to at least 3 features out of 5. Handedness similarity occurs the most frequently, followed by orientation, shape, movement, and position. However, similarity with respect to all form features is rare (only 3.82\% of gesture pairs). Section A of the Supplementary Materials provides further details on the inter-coder reliability of the annotation and the value distribution of each feature.

\subsection{Data Samples}
Using the manually segmented $4949$ gesture strokes as an anchor, we sample one-second windows around these strokes. We apply a sliding window with an offset of 2 frames and take any time windows that overlap $>50\%$ with gesture strokes. We do so because the segmented gesture strokes exclude other gesture phases, such as preparation, retraction and hold \cite{mcneill1992hand}, which can provide very useful information to represent gestures \cite{ghaleb2023co}. 
This procedure resulted in $70153$ gestural windows.
Since speech and co-speech gestures are semantically and temporally coordinated but not perfectly aligned, we take a larger speech window following ghaleb \etal~\cite{ghaleb2024le}. For each one-second gesture window, we include an extra half-second before and after, resulting in a two-second speech window.

For the speech modality, we use the underlying raw audio waveform of each speech time window. For the visual modality, we use MMPose \cite{sengupta2020mm} to obtain body poses from the gestural time windows. MMPose extracts the 2D positions (i.e., the $x$ and $y$ coordinates) of body joints and the detection confidence of 133 body joints. We utilize only 27 hands and upper body joints, which are the most relevant for sign language recognition \cite{jiang2021skeleton} and hand gesture detection \cite{ghaleb2023co}.
As proposed by Yan \etal~\cite{yan2018spatial}, we create spatio-temporal graphs from the detected poses. Each graph contains $j$ joints (vertices) and $e$ edges and is represented as $x = (j, e)$. The resulting graph has two types of edges: spatial edges that correspond to the natural connectivity of joints and temporal edges that link the same joints across successive frames. As a result, a time window of a skeleton is represented with a tensor as follows: $\boldsymbol{X}^G \in \mathbb{R}^{c \times t_v \times j}$, where $c$ is a joint data point ($x$, $y$, confidence), $t_v$ is the number of frames in a visual time window, and $j$ is the number of body joints.

\begin{figure*}[t]
    \centering
    \includegraphics[width=1\linewidth]{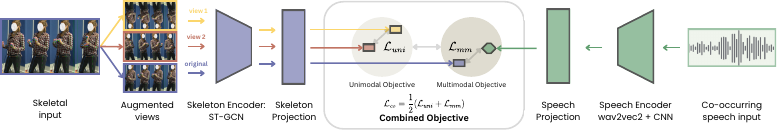}
    \caption{The proposed contrastive learning framework utilizing both unimodal and multimodal objectives.}
    \label{fig:cl_framework}
    \Description[]{}
\end{figure*}

\section{Models and Contrastive Objectives}
\label{sect:models}
In this study, we use contrastive self-supervised learning to refine the embeddings extracted by two pre-trained backbone (``tower'') models: a speech model (Section \ref{sec:wave2vec}) and a skeleton model (Section \ref{sec:gesture_backbone}).
Given that contrastive objectives aim to align representations, our goal here is to bring %
representations of speech and co-speech gestures closer in a latent space. 
More formally, given a dataset $\{\boldsymbol{x}_l^S, \boldsymbol{X}_l^G\}_{l=1}^N$ of size $N$ containing matching speech and gestural signals, their encoded representations can be written as $\{f_S(\boldsymbol{x}_l^S), f_G(\boldsymbol{X}_l^G)\}_{l=1}^N$, where $f_S(\cdot)$ and $f_G(\cdot)$ are feature encoders for the speech and gesture modality, respectively. We also use the projection heads $g_S(\cdot)$ and $g_G(\cdot)$ (Section \ref{sect:mlps_heads}), that map features from two modalities $g_S(f_S(\boldsymbol{x}_l^S)), g_G(f_G(\boldsymbol{X}_l^G))\}_{l=1}^N$ to the same size in a joint latent space \cite{chen2020simple, radford2021learning}. The proposed framework is illustrated in Figure \ref{fig:cl_framework}.

The following subsections describe the backbone models, projection heads, and contrastive learning objectives. Section B of the Supplementary Materials details the implementation and data augmentation methods. 
Additionally, all corresponding data and code needed to reproduce the results are publicly available on GitHub.\footnote{ \href{https://github.com/EsamGhaleb/Learning-Co-Speech-Gesture-Representations}{https://github.com/EsamGhaleb/Learning-Co-Speech-Gesture-Representations}}.

\subsection{Backbone Models and Projection Heads}
\label{sect:backbone_models}
\subsubsection{Speech Model}
\label{sec:wave2vec}
For the speech stream, we use the pre-trained \textsc{wav2vec}-2 model, which is trained with a masked language-modeling objective on large amounts of speech data \cite{baevski2020wav2vec}.
Concretely, we use the \texttt{wav2vec2-xlsr-300} version, 
pre-trained on the raw speech waveforms in multiple languages. The model first applies Convolutional Neural Networks (CNNs) to 1-D waveforms as a feature extraction layer, followed by $24$ Transformer encoders.
The CNN block extracts representations of fine-grained speech segments (20 milliseconds long). Sequences of these representations are then used as separate tokens in the 24 self-attention blocks. 
\textsc{wav2vec}-2 has shown state-of-the-art performance across different speech-based tasks, such as automatic speech recognition \cite{hsu2021hubert} and speech emotion recognition \cite{Chen2023}. 
We expect that the 
information related to speech prosody and content captured by this model will be beneficial for 
co-speech gesture representations. 

In our implementations, we build a downstream network on top of the output sequences corresponding to two-second speech windows, obtained from all 24 self-attention layers. We do not simply use the features of the last layer because outputs of different layers contain different amounts of semantic and prosodic information \cite{shah2021all, li2023exploration}, which can be influential for guiding co-speech gestures \cite{Nyatsanga2023}. As a result, the obtained representations have high dimensionality. To map information from these features into a lower-dimensional vector, we apply a weighted average with learnable weights for embeddings from each network layer. Furthermore, two pointwise CNN layers are implemented to fuse signals along the temporal dimension \cite{pepino2021emotion}. As a result, the output of the employed architecture is a one-dimensional vector containing $128$ dimensions.

\subsubsection{Skeletal Model}
\label{sec:gesture_backbone}
To encode the spatio-temporal graphs of body joints for each one-second gesture window, we use Spatio-Temporal Graph Convolutional Networks (ST-GCNs)  \cite{yan2018spatial}, which  
generalize traditional convolution operations to graph neural networks. 
ST-GCNs embed the bodily movements of each gestural time window, capturing their spatial-temporal dynamics. 
We use a pre-trained model for Sign Language Recognition (SLR) by Jiang \etal~\cite{jiang2021skeleton}. The model was trained for Turkish Sign Language recognition using only $27$ hand and upper body joints' coordinates. We refer to this model as the \textbf{\textit{SLR baseline model}}.
This model takes a spatio-temporal graph of a skeleton window as input and produces an embedding vector with $256$ dimensions.

\subsubsection{Projection Heads.}
\label{sect:mlps_heads}
The projection head for each modality consists of a three-layer MLP. The first layer is linear and has 128 dimensions for the speech head and 256 dimensions for the skeletal head. This layer is followed by a ReLU activation, which then leads to a final linear layer that projects both modalities into a 128-dimensional space. The objective of both heads is to project encoded features into a space that can be effectively compared through the contrastive learning method.

\subsection{Unimodal Contrastive Objective}
\label{sec:unimodal_objective}
We use the SimCLR \cite{chen2020simple} for training the encoder from the SLR baseline. This framework crafts the positive pairs for the contrastive objective by applying skeletal augmentations that generate multiple views of the same instance, training the model to build representations of the underlying gesture, regardless of the augmentations. We use a combination of mirroring, scaling, random moving, jittering, and shearing, each with a $50\%$ probability.
Thus, a gesture instance $\boldsymbol{X}_l^G$ is transformed into two views with two random skeleton augmentations $t_i(\boldsymbol{X}_l^G)$ and $t_j(\boldsymbol{X}_l^G)$. Then, the transformed views are passed through the encoder and projection to obtain the view representations $\boldsymbol{z}^G_i = g_G(f_G(t_i(\boldsymbol{X}_l^G)))$ and $\boldsymbol{z}^G_j = g_G(f_G(t_j(\boldsymbol{X}_l^G)))$ for the loss calculation. For this positive pair, the contrastive loss function treating $\boldsymbol{z}^G_i$ as an anchor can be written as follows:

\begin{equation}
    l_{uni}(i, j) = -log\frac{exp(\frac{s_(\boldsymbol{z}^G_i, \boldsymbol{z}^G_j)}{\tau})}{\sum_{k=1}^{2N_b} \mathbb{I}_{[k \neq i]} exp(\frac{s_(\boldsymbol{z}^G_i, \boldsymbol{z}^G_k)}{\tau})},
    \label{eq:NT-Xent}
\end{equation}
where $\tau$ is a temperature hyperparameter and $s(\cdot)$ is a cosine similarity, and $N_b$ is a batch size. The 
 \textit{\textbf{unimodal contrastive objective}} for the whole batch, then, is formulated as follows:
\begin{equation}
    \mathcal{L}_{uni} = \frac{1}{2N_b}\sum_{\{i, j\} \in N_b} {(l_{uni}(i, j) + l_{uni}(j, i))}.
    \label{eq:batch_loss_uni}
\end{equation}

\subsection{Multimodal Contrastive Objective}
\label{sec:cross_objective}
For multimodal training, we use Contrastive Multiview Coding \cite{tian2020contrastive}, also frequently referred to as the CLIP-like objective \cite{radford2021learning}.
This framework allows us to align and ground representations of co-speech gestures in corresponding speech without using data annotations. Concretely, contrastive learning aims to maximize the similarity between representations of matching gesture-speech pairs $\{\boldsymbol{z}^S_l = g_S(f_S(\boldsymbol{x}_l^S)), \boldsymbol{z}^G_l = g_G(f_G(\boldsymbol{X}_l^G))\}$ by contrasting them against other instances in mini-batches. In this case, the loss function for the $l$-th pair of instances contains two terms, $l_{mm}(G, S)$ and $l_{mm}(S, G)$. The former considers gestural features as anchors and mines positive and negative speech representations to form pairs, whereas the latter takes speech representations as anchors. More formally, the \textit{\textbf{multimodal contrastive objective}} $l_{mm}(G, S)$ can be written as follows:
\begin{equation}
    l_{mm}(G, S) = -log\frac{exp(\frac{s_(\boldsymbol{z}^G_l, \boldsymbol{z}^S_l)}{\tau})}{\sum_{k=1}^{N_b} exp(\frac{s_(\boldsymbol{z}^G_l, \boldsymbol{z}^S_k)}{\tau})},
    \label{eq:NT-Xent-mm}
\end{equation}

\noindent
Similarly to Equation \ref{eq:batch_loss_uni}, the loss for the whole batch with multimodal representations is formulated as follows:

\begin{equation}
    \mathcal{L}_{mm} = \frac{1}{2N_b}\sum_{l = 1}^{N_b} {(l_{mm}(G, S) + l_{mm}(S, G))}.
    \label{eq:batch_loss_mm}
\end{equation}

\subsection{Combined Objective} 
\label{sec:combined}
Finally, we propose an architecture that employs unimodal and multimodal objectives jointly to map related augmented skeleton representations closer while also grounding them in the co-occuring spoken language. We achieve this \textbf{\textit{combined objective}} as follows: $
    \mathcal{L}_{co} = \frac{1}{2} {(\mathcal{L}_{uni} +  \mathcal{L}_{mm}}).$

\section{Gesture Similarity Analysis}
\label{sect:results}
In this section, we first evaluate the representations obtained with our four models: the SLR baseline model and the three contrastive models with unimodal, multimodal, and combined objectives. For the three proposed contrastive models, the gestural representations are extracted from the last layer of the skeletal projection head. For the SLR baseline model, the representations are obtained from the last FC layer.
We then use the best model's gestural representations to study gesture similarity within and across speakers in face-to-face referential dialogue.

\subsection{Evaluation of Model Representations}
\label{sec:results_manual_coding}
We assess the effectiveness of the models' representations by examining the extent to which pairwise gesture similarity correlates with the number of form features shared by gesture pairs. This kind of \textit{intrinsic} evaluation of model representations against human similarity judgments is common in natural language processing \cite{navigli2019overview} and vision-and-language research \cite{bruni2014multimodal,pezzelle-etal-2021-tacl}. Here, we apply it for the first time to gesture representation learning. 

For this evaluation, we use the 419 gesture pairs manually coded for similarity in the dataset by Rasenberg~\etal~\cite{rasenberg2022primacy}. As explained in Section \ref{sect:manual_annot}, these gesture pairs have been annotated with 5 binary form features indicating whether the two gestures are similar or not with respect to handedness, shape, rotation, movement, and position. For each model, we compute the Spearman correlation coefficient between the pairwise cosine similarity of model representations and the number of form features these pairs share. The number of shared form features ranges from 0 (no similarity) to 5 (similar with respect to all features)---see Figure~\ref{fig:distribution_of_shared_feats}. Cosine similarity of model representations has a range of $[-1,1]$.

The SLR baseline model shows a correlation coefficient of $\rho=0.23$.
Surprisingly, the unimodal model trained with contrastive learning on our dataset gives a slightly lower correlation of $\rho=0.20$. 
In contrast, the multimodal model (i.e., the model trained with the multimodal contrastive objective) displays a correlation of $\rho=0.24$, 
i.e., slightly higher than the baseline. The highest correlation, $\rho=0.31$, is obtained with the representations learned by the model with the combined training objective. 
All correlation coefficients are significant, with $p <0.001$. 

\begin{figure}[ht!]
    \centering
    \includegraphics[width=0.9\linewidth]{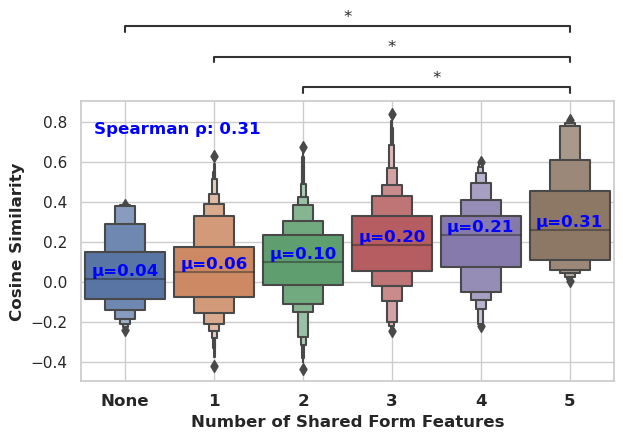}
    \caption{The cosine similarity scores distribution between pairs of gestures' representations, based on the number of shared form features in each pair. The similarity scores of gesture pairs sharing 5 form features are significantly higher than the similarity scores of gesture pairs sharing 2, 1 or 0 form features, respectively (t$>3.8$, p-value$<0.001$)}
    \label{fig:combined_model_and_manual_codings}
    \Description[]{}

\end{figure}

Figure \ref{fig:combined_model_and_manual_codings} shows the distribution of similarity scores by the best-performing model with the combined objective for gesture pairs that share different amounts of form features. As can be observed, when there are no shared features (i.e., zero similarity according to human judgments), the mean average similarity score is approximately $0.04$, significantly lower (according to an independent t-test with Bonferroni correction, t=$-3.855, p<0.001$) than the mean similarity when the two gestures share all form features, which is approximately $0.31$. The plots corresponding to the other models can be found in Section C.1 of the Supplementary Materials.

In sum, we conclude that the model with the combined contrastive objective is able to learn robust gesture representations that are well aligned with manually coded representations to a larger extent than the representations learned by the other models we test. In Section~\ref{sec:probing_analysis}, we will dive deeper into the model representations to investigate whether they encode specific features.

\subsection{Gesture Similarity in Referential Dialogues}
\label{sec:sim-analyses}
In this section, we investigate to what extent the learned gesture representations comply with well-motivated expectations regarding gesture similarity within and across speakers. We report the results obtained with the best model's representations according to the evaluation carried out above, but the same pattern of results holds for the four models we test (see Section C.2 in the Supplementary Material).
For these analyses, we include the model representations for all gestures in the dataset by Rasenberg~\etal~\cite{rasenberg2022primacy}. 
Recall that all gestures are annotated with their referent.

\subsubsection{Referent vs.~Speaker Driven Similarity}
\label{sec:referent-similarity}
In the first analysis, we focus on gestures that take place within each two-participant dialogue and investigate the interplay between two variables regarding gesture pairs: the \textit{speaker} (whether two gestures are produced by the same speaker or not) and the \textit{referent} (whether two gestures refer to the same object or not). We formulate the following hypotheses:

\begin{itemize}[leftmargin=24pt,itemsep=3pt]
\item[H1 a.] Representations of gestures by the same speaker will be more similar if the gestures have the same referent than if they refer to different objects.
\item[\ \ \ b.] Representations of gestures made by different speakers will be more similar if the gestures have the same referent than if they refer to different objects.
\item[H2 \; ] Representations of gestures with the same referent will be more similar if the gestures are produced by the same speaker than if they are made by different speakers.
\end{itemize}

\noindent
The two branches of hypotheses H1 are motivated by the iconicity of the gestures present in the dataset we analyze: iconic gestures resemble what they depict; in addition, the use of an iconic gesture is often concurrent with a referential expression articulated in speech. 
For example, to refer to the object part highlighted in Figure~\ref{fig:similarity_example}, the participants produce a gesture resembling a book while uttering \textit{``you mean with that open book, so to speak''}, \textit{``yeah a bit like you've opened a book''}.
Hence, we hypothesize that our model representations should be well-equipped to capture the visual and speech similarities that tend to characterize same-referent gestures, both when they are produced by the same speaker (H1a) as well as when they are produced by different speakers (H1b). 

Hypothesis H2 is motivated by the assumption that there are individual idiosyncrasies in how speakers gesture and speak. For example, the relationship between systematic and idiosyncratic patterns in the production of iconic gestures has been studied by Bergmann and Kopp \cite{bergmann2010systematicity, bergmann2009increasing}, where both factors were found to impact gesture formation. Systematic factors and dialogue context affect the production of iconic gestures across speakers, but individuals may differ in how they realize their gestures, e.g., regarding the amount of shaping or drawing motions they use.

To test whether the learned representations comply with these hypotheses, we extract four sets of gesture representations: 
\textit{same-referent-same-speaker} (7K pairs), 
\textit{same-referent-different-speaker} (5.5K pairs), 
\textit{different-referent-same-speaker} (455K pairs), and 
\textit{different-referent-different-speaker} (358K pairs).\footnote{We use the total number of pairs available for each of these sets; the results we present next and their statistical significance remain stable when the number of pairs is downsampled (e.g., to 5.5k for each set).}

\begin{figure}[ht!]
    \centering
    \includegraphics[width=0.9\linewidth]{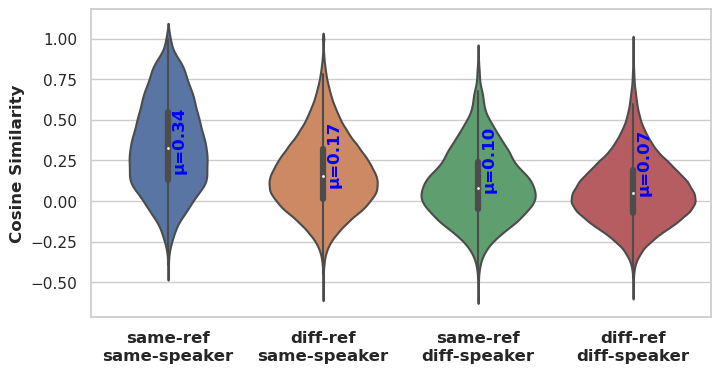}
    \caption{Distribution of cosine similarity scores of self and across-speaker gesture pairs in a dialogue when referring to the same or different referents. The labels `same-ref' / `diff-ref' indicate whether the gestures in a pair refer to the same object or not. According to the independent t-test with Bonferroni correction, distributions of similarity scores in all sets are significantly different.}
    \label{fig:same_spaker_same_vs_diff_objects}
    \Description[]{}

\end{figure}

\paragraph{Results}
Figure \ref{fig:same_spaker_same_vs_diff_objects} shows the distribution of cosine similarity scores for each of these sets of gesture pairs. We observe that gesture pairs produced by the same speaker to refer to the same referent exhibit the highest degree of similarity ($\mu=0.34$), while gesture pairs produced by different speakers to refer to different objects are the least similar ($\mu=0.07$). 
In compliance with H1a, gestures by the same speaker have lower similarity when they are not referentially related ($\mu=0.34$ vs.~$\mu=0.17$). The same trend holds for gestures by different speakers, in line with H1b: their similarity is higher when the gestures refer to the same object ($\mu=0.10$ vs.~$\mu=0.07$). Finally, we find that gestures with the same referent are more similar when they are made by the same speaker than by different speakers ($\mu=0.34$ vs.~$\mu=0.10$), which complies with H2. All mean differences are significantly different according to an independent t-test with Bonferroni correction. 

In conclusion, the similarity of gesture pair representations learned by the models %
follows expected patterns that reflect both the interplay of speech and vision in iconic gestures that are referentially related, as well as individual characteristics of speakers.

\subsubsection{Referent vs.~Interaction Driven Similarity}
\label{sec:interaction-similarity}
In the previous analysis, we observed that referentially related gestures by different speakers are significantly more similar than cross-speaker gestures that do not refer to the same object ($\mu=0.10$ vs.~$\mu=0.07$ in Figure \ref{fig:same_spaker_same_vs_diff_objects}). 
Is the similarity between referentially related gestures across speakers exclusively due to the shared referent (i.e., to the iconic character of the gestures), or is it also influenced by dialogue participants synchronizing their gestures through interaction? In studies of face-to-face dialogue, there is evidence that interlocutors tend to mimic each other's behavior---regarding both speech and gestures---when interacting \cite{holler2011co, rasenberg2022primacy, Akamine2024sp, ghaleb2024an}. We therefore formulate the following hypothesis:

\begin{itemize}[leftmargin=20pt]
    \item[H3 \ ] Representations of gestures by different speakers will be more similar when the two speakers are interlocutors within a dialogue than when the speakers are from different dialogues.
\end{itemize}

\noindent
To test whether the learned model representations confirm this hypothesis, we consider the two sets of gesture pairs by different speakers from the same dialogue we had extracted for the previous analysis (\textit{same-referent-different-speaker} and \textit{different-referent-different-speaker}) and extract two additional sets of pairs from different dialogues: \textit{same-referent-different-speaker-diff-dialogue} (137K pairs) and \textit{different-referent-different-speaker-diff-dialogue} (10M pairs). 

\begin{figure}
    \centering
    \includegraphics[width=0.9\linewidth]{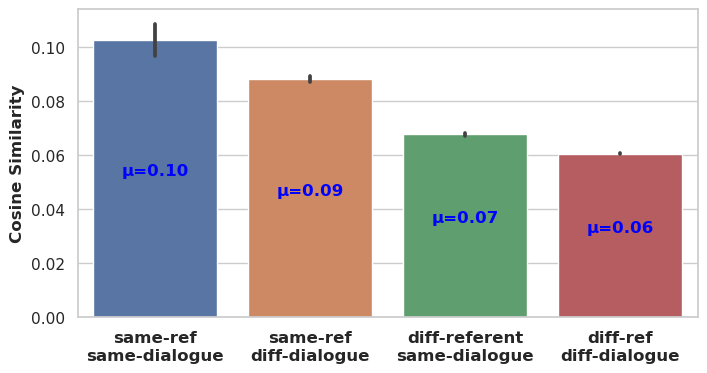}
    \caption{Similarity scores of gesture representation pairs by different speakers. The labels `same-ref' / `diff-ref' indicate whether the gestures in a pair refer to the same object or not; the labels `same-dialogue' / `diff-dialogue' indicate whether the speakers are interlocutors within the same dialogue or are participants from different dialogues. An independent t-test with Bonferroni correction shows that all sets' distributions of similarity scores are significantly different.}
    \label{fig:Cross_and_different_speakers}
    \Description[]{}
\end{figure}

\paragraph{Results} 
Figure \ref{fig:Cross_and_different_speakers} shows the mean similarities for the sets of pairs mentioned above. 
The results show that referentially related gesture pairs are most similar if they are produced by speakers interacting within a dialogue rather than by speakers who refer to the same object but do not directly interact with each other. 
Furthermore, even for gestures that do not refer to the same object, the representations across speakers are more similar when they are interlocutors in a dialogue. Taken together, these results lend evidence to H3: the model representations reflect that gesture similarity is partially driven by alignment processes due to dialogic interaction.

\section{Probing Analysis}
\label{sec:probing_analysis}

The analysis presented in Section \ref{sec:results_manual_coding} showed that the similarity between model gesture representations significantly correlates with manually coded gesture similarity. However, that analysis does not reveal to what extent the latent representations may encode interpretable features. In this section, we shed light on this issue via diagnostic probing \cite{belinkov2022probing}, a technique widely applied to decode linguistic properties present in textual representations \cite{belinkov-etal-2017-neural, conneau-etal-2018-cram, ryb-etal-2022-analog} that has also been used to interpret features of other modalities, including visual signals \cite{basaj2021visualprobing} and spoken utterances \cite{de2022probing}.

Diagnostic probing is frequently formulated as a classification task. Specifically, a classifier is trained on top of the learned representations to predict a certain property. Here, we use the 419 gesture pairs manually annotated with 5 binary features and build 5 classification models, one per feature. For each pair of representations, the probing classifiers predict whether the representations are similar in handedness, shape, rotation, movement, or position.

\subsection{Probing Models}
We probe the representations learned by the model with the combined contrastive objective, in particular, the representation extracted from the last layer of the ST-GCN encoder. For comparison, we also probe the representations of the baseline SLR-skeleton model, taken from the same layer.  

\paragraph{Probe architecture.}
Our probing models need to operate on top of pairs of representations as the binary labels are provided for pairs of gestures. Given that the gestures in each pair are by different speakers and there may be contrasts in input skeletal joints due to different camera viewpoints (e.g., mirrored views),  
we pass each representation in a pair through a simple linear layer with 32 neurons, followed by ReLU activation. 
Subsequently, the projected vectors are concatenated and used as input to a linear probe with sigmoid activation. Finally, the binary cross-entropy loss function is computed between the output of the probing model and the ground truth binary label indicating the similarity of gestures. 

\paragraph{Random baseline.}
While the idea of diagnostic probing is simple, interpreting the results of a probing classifier requires careful consideration, as probing models may capture irrelevant patterns \cite{belinkov2022probing}. To account for this, we also train our 
probing model using random features. Specifically, these features are generated by an encoder following the same ST-GCN architecture but initialized with random weights. If the performance of the probe with the actual gesture representations is significantly higher than its performance with random features, we can conclude that the model representations encode, to some extent, the property being probed for. 

\paragraph{Implementation details.} The model is trained for 50 epochs using the Adam optimizer with a learning rate of $5e^{-4}$ for parameter tuning. For each property and type of gesture representation, 100 experiments were conducted using train, validation and test splits 
sampled with different random seeds and containing 60, 20 and 20\% of gesture pairs, respectively. The probing models are evaluated using the ROC-AUC metric on the test set. For measuring the statistical significance of the observed differences, we employ the Mann-Whitney U-test ($\alpha=0.05$) and apply Benjamini-Hochberg correction for multiple testing.
\begin{figure}[t]
    \centering
    \includegraphics[width=1\linewidth]{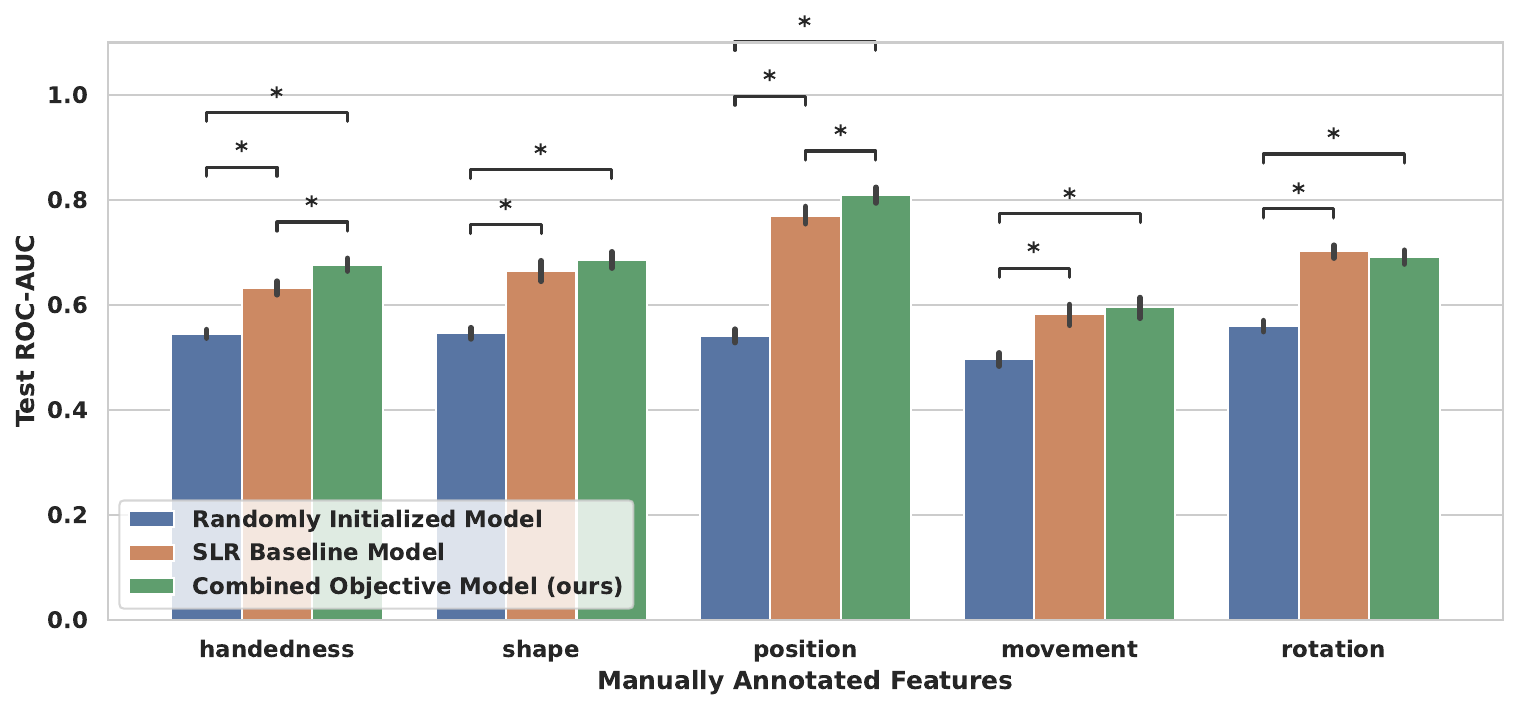}
    \caption{Test ROC-AUC scores of the probing models for each type of gestural similarity feature. The statistically significant differences are highlighted with * when observed.}
    \label{fig:probing_results}
    \Description[]{}
\end{figure}
\subsection{Results}

The results of the probing analysis are summarized in Figure \ref{fig:probing_results}. The first result that stands out is that, for all the manually annotated features, the probes operating on model representations (by both the SLR model and the combined objective model) perform significantly better than those operating on randomly initilized representations. This suggests that information about all these properties---handedness, shape, position, movement, and rotation---is somewhat encoded by the models. Yet, the relative increase with respect to the random baseline differs per property, with position exhibiting the largest increase (more than $0.25$ in mean ROC-AUC) and movement the smallest ($0.1$ mean ROC-AUC). In addition, we observe that the combined objective model representations tend to yield higher results than those by the SLR model for all features except rotation. Concretely, this advantage is statistically significant for handedness and position. This suggests that speech may be more connected to these features of gestures when they are represented through the skeletal joints.

In sum, our analysis indicates that the models, in particular our combined objective model, learn gesture representations that encode different properties identified by experts through manual annotation. The results also suggest that information about movement appears to be the most challenging for the models to learn.

\section{Conclusion}
\label{sect:conclusion}

We proposed an approach for learning co-speech gesture representations using contrastive self-supervised learning. This approach employs loss functions to bring two views of gestural movements closer in a latent space while also grounding them in spoken language. To train the proposed models, we used segmented gestures obtained from a dataset of face-to-face referential dialogues without any additional labeling.
While most work on gesture representation learning has focused on task-specific evaluation, in this paper, we performed a thorough intrinsic evaluation of the learned representations by analyzing the extent to which they exhibit desirable properties that align with expert intuitions. 
Our results showed that a model that combines both unimodal and multimodal contrastive objectives has advantages over a Sign Language Recognition baseline model and other contrastive variants. Concretely, we observed that (1) the representations learned by the combined objective model achieve the highest correlation with pair-wise similarity of form features annotated by experts, and (2) these form features, particularly handedness and position, can more accurately be decoded from the latent representations learned by this model, as demonstrated by a probing analysis. 
This analysis also suggests that information about gesture movement is challenging for the models to learn. Future work could address this by encoding architectures that can exploit additional aspects of skeletal data, such as bones and motions.

We also examined how different factors, such as gesture iconicity, individual speaker characteristics, and dialogue coordination, influence the similarity between gesture pairs. Our analyses showed that model representations comply with theoretically-motivated expectations regarding these factors. In particular, they appear to capture three major sources of variance: degree of (1) idiosyncratic personal style, (2) common ground building in conversation, and (3) conventional iconic gesture use. We believe that this makes them a valuable tool for gesture analysis studies. 
When analyzing the data in more detail, beyond the hypotheses investigated in Section~\ref{sec:sim-analyses}, we observed that for the gesture pairs where referents are different and the speaker is the same, the similarity of the representations learned by the multimodal models is higher than when the referent is the same, and the speakers are different (more details are available in Section C.2 of the Supplementary Materials). This suggests that the inclusion of speech information during training may amplify the encoding of individual speaker differences. We leave an in-depth investigation of this aspect to future work.

\begin{acks}
This research was funded by the Dutch Research Council (NWO) under a Gravitation grant (024.001.006) awarded to the Language in Interaction consortium.
EG and RG are supported by the European Research Council (ERC) grant agreement number 819455.
\end{acks}

\bibliographystyle{ACM-Reference-Format}
\bibliography{main}

\end{document}